\documentclass[11pt]{article}

\usepackage[preprint]{acl}

\usepackage{times}
\usepackage{latexsym}
\usepackage{amssymb} 
\usepackage{booktabs}
\usepackage[ruled,vlined,linesnumbered]{algorithm2e}
\usepackage[table]{xcolor}
\usepackage{multirow}
\usepackage{placeins}
\usepackage{float}
\usepackage{listings}
\usepackage{graphicx}
\usepackage{enumitem} 
\usepackage{setspace} 
\usepackage{tabularx}
\usepackage{adjustbox}
\usepackage{xcolor}
\usepackage{makecell}
\usepackage{ragged2e} 
\usepackage{listings}

\usepackage{tikz}
\usetikzlibrary{shapes, arrows.meta, positioning, fit}

\usepackage[most]{tcolorbox}
\tcbuselibrary{listings, skins, breakable}
\usepackage{caption}
\usepackage{dblfloatfix}  

\newtcolorbox{promptbox}[1][]{
  enhanced,
  breakable=false,
  colback=gray!3,
  colframe=black!80,
  coltitle=white,
  colbacktitle=black!80,
  title=#1,
  fonttitle=\bfseries\normalsize,
  fontupper=\footnotesize, 
  arc=2pt,
  outer arc=0pt,
  boxrule=0.8pt,
  top=2pt,
  bottom=2pt,
  left=5pt,
  right=5pt,
  boxsep=0pt,
  middle=1pt
}

\usepackage[T1]{fontenc}

\usepackage[utf8]{inputenc}

\usepackage{microtype}

\usepackage{inconsolata}

\usepackage{graphicx}
\usepackage{amsmath}
\usepackage{cleveref}
\usepackage{subcaption}

\graphicspath{{images/} }

\title{Graph-Based Chain-of-Thought Pruning for Reducing Redundant Reflections in Reasoning LLMs}

\author{
 \textbf{Hongyuan Yuan\textsuperscript{1,2}},
 \textbf{Xinran He\textsuperscript{2}},
 \textbf{Run Shao\textsuperscript{1,2}},
 \textbf{Bolei He\textsuperscript{2}},
\\
 \textbf{Xianwei Xue\textsuperscript{2}},
 \textbf{Mengke Chen\textsuperscript{2}},
 \textbf{Qiutong Pan\textsuperscript{2}},
 \textbf{Haiwei Wang \textsuperscript{2}},
 \textbf{Haifeng Li\textsuperscript{1,*}}
\\
\\
 \textsuperscript{1}School of Geosciences and Info-Physics, Central South University, Changsha, China,
 \\
 \textsuperscript{2}Baidu Inc., Beijing, China,
\\
\texttt{\{hexinran, hebolei, xuexianwei, chenmengke, panqiutong, wanghaiwei\}@baidu.com}, \\
\texttt{\{yuanhongyuan, shaorun, lihaifeng\}@csu.edu.cn}
}

\begin{document}
\maketitle
\begin{abstract}

Extending CoT through RL has been widely used to enhance the reasoning capabilities of LLMs. However, due to the sparsity of reward signals, it can also induce undesirable thinking patterns such as overthinking, i.e., generating redundant intermediate reasoning content.
In this work, we argue that a major source of such redundancy is inefficient reflection, which often manifests in two problematic patterns: Indiscriminate Reflection, where the model performs broad, low-impact checks throughout reasoning, and Repetitive Reflection, where it repeatedly re-verifies an already established conclusion.
To address this, we introduce a graph-based CoT optimization framework.
Specifically, we convert each linear CoT into a directed acyclic graph (DAG) with explicit dependency edges, and design a dual pruning strategy: branch-level pruning removes weakly contributing reflection branches, while depth-level pruning eliminates late-stage re-verification.
We distill this behavior via a three-stage pipeline: (1) SFT to initialize the policy on pruned concise traces, (2) DPO to prefer correct but less redundant trajectories, and (3) GRPO with length penalty to jointly optimize answer correctness and efficiency. Experiments show that our approach reduces the average reasoning tokens by 42\% while maintaining or improving accuracy.

\end{abstract}

\section{Introduction}

Recently, test-time scaling has emerged as an important pathway for improving the reasoning capabilities of LLMs. Representative works such as o1\cite{openaio1} and R1\cite{deepseekr1} employ reinforcement learning with verifiable rewards to encourage the generation of longer CoT, leading to substantial gains on challenging tasks such as mathematics, code, and logical reasoning. However, reward signals in RL training are often sparse and delayed, making credit assignment over long trajectories more difficult\cite{tran2025exploitingtreestructurecredit}; as a result, models may develop inefficient thinking patterns such as overthinking, namely producing a large amount of redundant intermediate reasoning that contributes little to the final answer and increases inference cost\cite{donotthinkthatmuch, sui2025stopoverthinkingsurveyefficient}.

\begin{figure}[t]
    \centering
    \includegraphics[width=\columnwidth]{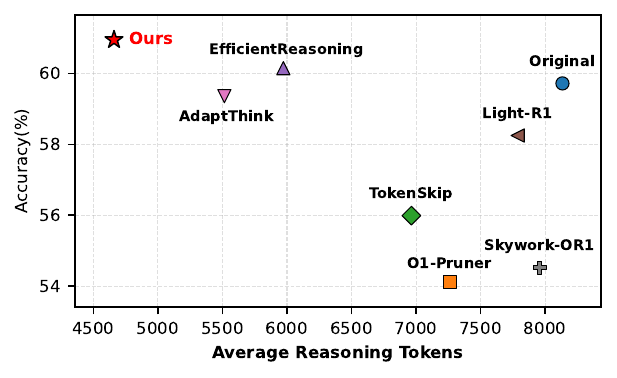}
    \caption{Relationship between average reasoning tokens and accuracy averaged across all benchmarks for different methods.}
    \label{fig:tokens_vs_aime25_accuracy}
\end{figure}

To mitigate this issue, prior work has explored different ways to constrain and suppress redundancy in CoT. One line of methods focuses on length and reasoning budget control, with typical examples like TALE\cite{han2025tokenbudgetawarellmreasoning} and CoT-Valve\cite{cot-valve}. Another stream focuses on detecting and removing redundant content within the generated trajectory, for example, TokenSkip\cite{xia-etal-2025-tokenskip} performs controllable compression based on token-level semantic importance, while Think Clearly\cite{choi2025thinkclearlyimprovingreasoning} and TRAAC\cite{singh2025thinkrightlearningmitigate} leverages attention-based criteria to detect redundant tokens.

In this work, we argue that redundancy largely stems from inefficient reflection behavior. We identify two major forms: Indiscriminate Reflection, where the model checks every intermediate step even when the step is trivial; and Repetitive Reflection, where the model repeatedly re-checks intermediate conclusions that have already been verified. The common issue is that such reflection does not contribute new useful information to the main reasoning chain. To precisely identify redundant reflection, we need to recover the dependency relations among reasoning steps. Since practical reasoning often involves non-linear behaviors such as branching exploration, backtracking, and cross-step repetition, we propose to explicitly model CoT with a graph structure, representing reasoning units and their dependencies as nodes and edges, and identifying redundant reflection based on graph properties.

Specifically, we first split the raw CoT into a sequence of steps using keyword triggers, and then perform iterative graph construction with a LLM. With a carefully designed prompt, the model processes each step in sequence, and conditioned on the partially constructed graph and the current step, it creates semantically complete and functionally atomic nodes, determines their semantic roles (Progress nodes that advance reasoning or Review nodes that verify intermediate states) and their logical predecessors, and finally reconstructs the linear text into a DAG with explicit dependencies.

On top of this topology, we design a dual pruning strategy to precisely remove two types of redundancy. For Indiscriminate Reflection, we apply branch-level pruning: when a Review node produces only a small number of child nodes, it forms a narrow side branch that is unlikely to develop into the main reasoning chain and contributes little to the overall solution, and thus is removed. For Repetitive Reflection, we apply depth-level pruning: when a Review node appears in the later part of the CoT, it often corresponds to repeatedly re-verifying conclusions that have already been validated, which is unlikely to introduce new useful information; such nodes are treated as redundant and are removed.

During training, we first perform cold-start supervised fine-tuning on the pruned high-quality CoTs, so that the model initially acquires an efficient reasoning paradigm. To further internalize this strategy, we adopt a two-stage reinforcement learning procedure. In the first stage, we use DPO for preference alignment by constructing preference pairs (concise trajectories versus redundant trajectories), encouraging the model to reduce the probability of generating inefficient reasoning steps. In the second stage, we introduce GRPO with length penalty, using final-answer correctness as the primary signal while jointly optimizing reasoning efficiency, thereby further compressing redundant steps at inference time and improving robustness and accuracy on complex reasoning tasks. Experiments show that compared with strong baselines, our method can substantially reduce ineffective reasoning overhead while maintaining or improving reasoning accuracy.

\section{Related Work}

\subsection{Large Reasoning Models}

To elicit human-like step-by-step reasoning, researchers proposed Chain-of-Thought (CoT) prompting, which substantially improves LLM performance on complex reasoning tasks~\citep{wei2023chainofthoughtpromptingelicitsreasoning,kojima2023largelanguagemodelszeroshot}. Later work attempted to internalize this capability via SFT on CoT-annotated data, enabling models to naturally generate intermediate reasoning steps at inference time~\cite{shridhar-etal-2023-distilling,liao-etal-2025-skintern}. However, further gains are often limited by the scale and quality of available CoT datasets.

Recently, models such as~\cite{openaio1,deepseekr1} have shifted toward Reinforcement Learning, which uses outcome-based supervision in domains with verifiable answers (e.g., mathematics and code) and has proven effective for long-chain reasoning~\cite{wen2025reinforcementlearningverifiablerewards,su2025crossingrewardbridgeexpanding}. Along this direction, post-training with reinforcement learning has become increasingly central, with Group Relative Policy Optimization (GRPO)~\cite{shao2024deepseekmathpushinglimitsmathematical} emerging as a widely adopted recipe for improving reasoning performance.

\subsection{Efficient Reasoning}

Although longer CoT can improve LLM reasoning, it may also induce overthinking, i.e., generating redundant reasoning content~\cite{sui2025stopoverthinkingsurveyefficient}. 
To mitigate this issue, existing methods can be broadly categorized. The first stream directly regulate the overall reasoning budget by predicting/assigning a token budget\cite{han2025tokenbudgetawarellmreasoning,cot-valve} or introducing length-aware training objectives in reinforcement learning~\cite{luo2025o1prunerlengthharmonizingfinetuningo1like,arora2025traininglanguagemodelsreason}. Another stream involves identifying redundancy within a trajectory, for instance, Think Clearly~\cite{choi2025thinkclearlyimprovingreasoning} and TRAAC\cite{singh2025thinkrightlearningmitigate} leverages attention-based criteria to detect redundant tokens, TokenSkip~\cite{xia-etal-2025-tokenskip} relies on explicit importance scoring, while SPIRIT and Step-Entropy~\cite{cui-etal-2025-stepwise,li2025compressingchainofthoughtllmsstep} use perplexity-/entropy-related cues to identify and remove low-information steps/tokens. In contrast to the above methods, we instead focus on the reflective behaviors of the model, and directly reduce redundant self-reflection steps in the reasoning trace.

\section{Methodology}

\begin{figure*}[ht]
    \centering
    \includegraphics[width=\textwidth]{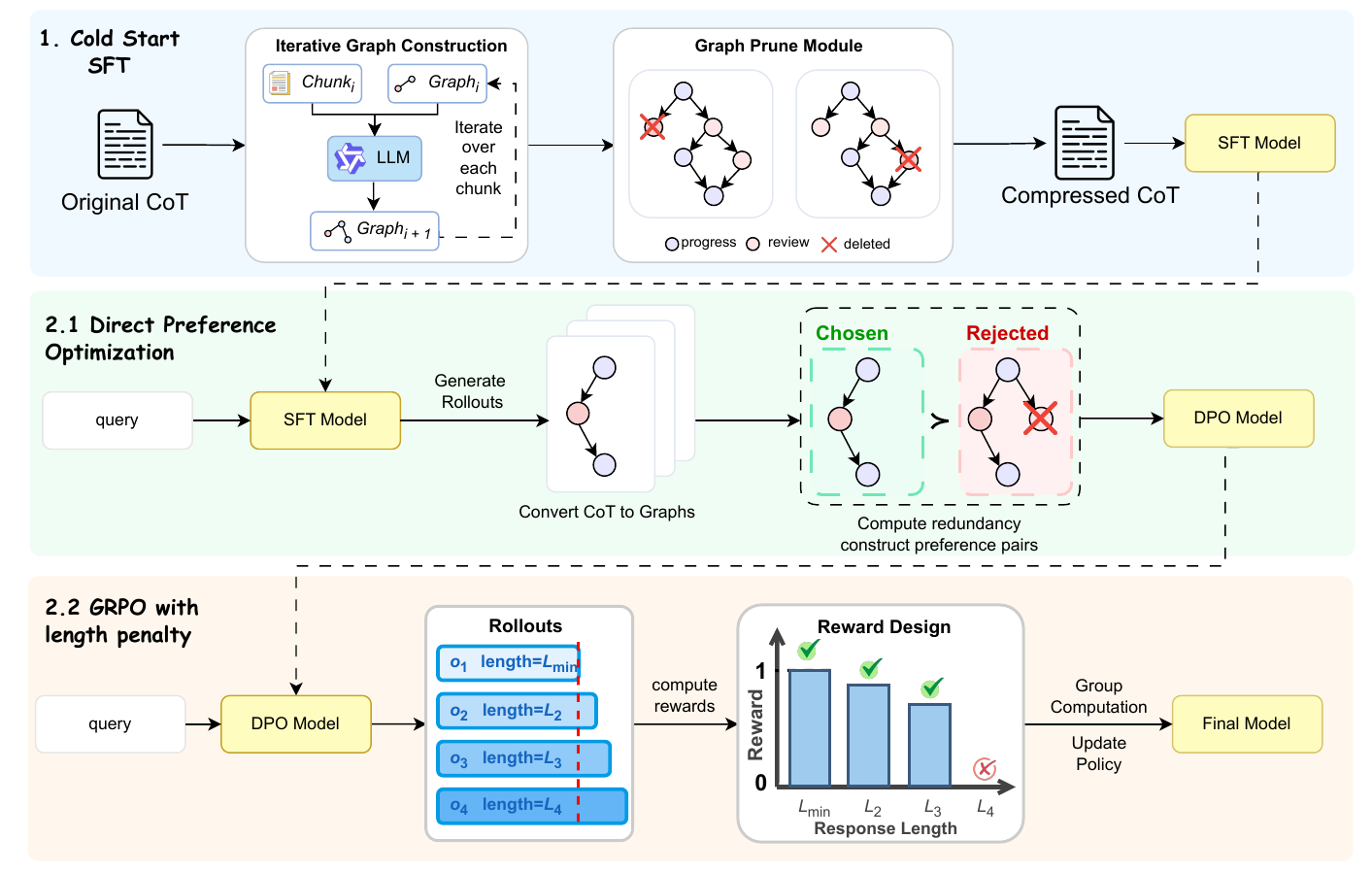}
    \caption{Training pipeline of our graph-based redundancy-aware post-training framework.
We first perform cold-start SFT using compressed chain-of-thoughts obtained via iterative graph construction and redundancy pruning, which instills an inductive bias toward concise reasoning. On top of the SFT model, we conduct DPO by ranking correct trajectories according to redundancy, preferring low-redundancy outputs over highly redundant ones. Finally, we apply GRPO with length penalty, jointly optimizing answer correctness and reasoning efficiency to obtain the final model.}
    \label{fig:overview}
\end{figure*}

\subsection{Problem Definition}

Formally, let $M_{\theta}$ be a large reasoning model. For a given problem $q$, the model generates its reasoning $r$ and produces a final answer $a$: $r, a\sim M_{\theta}(q)$, where the reasoning part can be split into many chunks $r=[c_1,c_2,...c_n]$ at special control tokens such as “wait”, “alternative”, etc.(complete list is shown in Appendix~\ref{app:split_tokens}). Each chunk represents an individual step generated by the model and $n$ denotes the total number of reasoning chunks.

Our objective is to propose a method to locate and remove redundant reflective segments in $r$:
\[
\tilde{r} \;=\; \mathrm{Prune}(r) \subseteq r,\quad \tilde{r} = [\tilde{c}_1,\ldots,\tilde{c}_{\tilde n}],
\]
and finetune the model $M_{\theta}$ using the dataset $D=\{(q,\tilde{r},a)\}$ so that the finetuned model learns to generate more efficient and concise reasoning paths at inference time, while maintaining or even enhancing its overall task performance and accuracy.

\subsection{Graph Construction and Pruning}
\subsubsection{Formatting Chain-of-Thought into a Graph}

Given a problem $q$ and its corresponding CoT $r=[c_1,\ldots,c_n]$, we progressively convert the linear reasoning sequence into a DAG
\[
G_t=(V_t,E_t,\ell_t),\quad t=0,1,\ldots,n,
\]
where $V_t$ is the node set, $E_t\subseteq V_t\times V_t$ is the edge set encoding logical dependencies between nodes, and $\ell_t:V_t\to\{progress,review\}$ is the node-type labeling:
\begin{itemize}
    \item \textbf{progress}: advances the active reasoning frontier, producing information that later steps rely on.
    \item \textbf{review}: reads, checks, restates, deletes, or rewinds existing material without advancing the frontier.
\end{itemize}

We initialize the graph as $G_0=(\varnothing,\varnothing,\varnothing)$ and update it sequentially with each chunk $c_i$ where $c_i$ is a step-level chunk obtained by splitting the CoT. At step $i$, conditioned on the current graph $G_{i-1}$ and the chunk $c_i$, we prompt an external LLM (e.g., qwen-turbo) to predict an operation:
\begin{equation}
o_i \sim f_{L}(G_{i-1},c_i)
\end{equation}
where $f_L$ is the prompting process and $o_i \in \{insert,merge\}$ indicate where to insert a new node or merge the current chunk into an existing node. Detailed prompts are provided in Appendix~\ref{app:graph_prompts}

When $o_i=insert$, the model will create a new node $v_i=(s_i,l_i)$, where $s_i$ is the summarization of the node, and $l_i$ is the type of the node, and then connect $v_i$ to existing nodes based on their dependencies. When $o_i=merge$, the model will select a target node $v^\star\in S_i$ and update its content based on the content of the current chunk.

To validate the constructed graphs, we further conduct a human evaluation on randomly sampled nodes, assessing both node-type labeling accuracy and node-level semantic atomicity. Detailed evaluation protocols and quantitative results are reported in Appendix~\ref{app:graph_eval}.

\subsubsection{Graph-Based Pruning Criteria}

After formatting the linear CoT into a DAG $G=(V,E,\ell)$, we aim to identify and prune low-utility \texttt{review} nodes, yielding a more compact and compute-efficient reasoning trajectory. 

We define the (directed) descendant set of a node $v$ as
\[
\mathrm{Desc}(v)\;=\;\{\,u\in V \mid v \leadsto u,\ u\neq v\,\},
\]
i.e., all nodes reachable from $v$ (excluding $v$ itself). The descendant count is
\[
B(v)\;=\;|\mathrm{Desc}(v)|.
\]
We also define the depth of $v$, denoted $d(v)$, is the length of a shortest path from the source node (in-degree $0$) to $v$, and we define the maximum depth $d_{\max}= d(t)$ where $t$ is the terminal node.

We define two types of redundancy in review nodes and prune nodes that fall into either category:
\noindent\textbf{Branch-Level Redundancy.}
A review node is considered redundant when it has fewer than $k$ descendants, i.e., $B(v) < k$. In this case, the node initiates only a narrowly expanding side branch that rarely develops into the main reasoning trajectory. Such nodes contribute little to the global problem-solving process and are therefore pruned.

\noindent\textbf{Depth-Level Redundancy.}
LLMs often reach the correct answer at an intermediate depth but continue producing additional rounds of reflection or self-checking afterward. We treat a review node as redundant if it appears in the late part of the trajectory, i.e., its relative depth exceeds a threshold $m$, $\frac{d(v)}{d_{\max}} > m$. Such nodes typically correspond to post-answer backtracking and add no new information to the reasoning chain, and are thus removed.

Finally, we locate and prune redundant review nodes, then relinearize the remaining DAG into a CoT sequence for training; we set $m=0.9$ and $k=2$ throughout all experiments.

\subsection{Training Pipeline}

To effectively suppress redundant reflection while preserving correct reasoning, we adopt a three-stage training pipeline for the policy model $M_\theta$: 
(1) cold-start supervised fine-tuning, which provides a cold-start initialization by training the model to follow concise reasoning traces; 
(2) preference-based optimization, which aligns the policy toward correct reasoning trajectories with lower reflective redundancy; and 
(3) GRPO with length penalty, which further refines the policy by jointly optimizing answer correctness and reasoning efficiency. The overall workflow is summarized in Algorithm~\ref{alg:three_stage}.

\subsubsection{Cold-Start Supervised Fine-Tuning}

In this stage, we instill an explicit inductive bias toward concise, non-redundant reasoning, which is essential for effective preference optimization and reinforcement learning in later stages. Starting from raw CoT annotations, we construct reasoning graphs and prune reflective redundancies to obtain concise reasoning traces $\tilde r$. 
For each problem $x$, we pair the pruned trace $\tilde r$ with its corresponding final answer $a$, and train the policy to generate the combined output sequence $y=(\tilde r,a)$. 
The supervised fine-tuning objective is the standard next-token negative log-likelihood:
\[
\mathcal{L}_{\text{SFT}}(\theta)
= -\mathbb{E}_{(x,\tilde r,a)} \sum_t \log \pi_\theta(y_t\mid x,y_{<t}),
\]
where $\pi_\theta$ is the policy model parameterized by $\theta$, $y_t$ denotes the $t$-th token in $y$, and $y_{<t}$ denotes the preceding token sequence.

\subsubsection{Two-stage Reinforcement Learning}

\paragraph{Stage I: Preference Optimization over Reasoning Trajectories.}

To further align the policy with concise reasoning behavior, we perform DPO on trajectory pairs ranked by redundancy. For each question $x$, we sample multiple trajectories and compute a redundancy score:
\[
R(y)=\frac{|\{v\in V:\ell(v)=\texttt{review}\}|}{|V|}+\frac{|y|}{\overline{|y|}_x},
\]
where $|y|$ is the number of generated tokens and $\overline{|y|}_x$ is the mean length of sampled trajectories for the same question $x$;
Among trajectories that yield correct final answers, we treat those with lower redundancy as preferred $y^+$ and those with heavier redundancy as dispreferred $y^-$. We then train the policy model with DPO to increase the relative likelihood of concise, low-redundancy trajectories. The DPO loss is
\begin{align*}
\mathcal{L}_{\text{DPO}}(\theta)
&= -\mathbb{E}_{(x,y^+,y^-)}
   \log \sigma\Big(
   \beta\big[
     \log\frac{\pi_\theta(y^+\mid x)}{\pi_{\text{ref}}(y^+\mid x)} \\
&\qquad\qquad
     - \log\frac{\pi_\theta(y^-\mid x)}{\pi_{\text{ref}}(y^-\mid x)}
   \big]\Big).
\end{align*}
Here $\pi_\theta$ is the current policy, $\pi_{\text{ref}}$ is a fixed reference policy (e.g., the SFT model), and $\beta>0$ controls the sharpness of the preference. 

\paragraph{Stage II: GRPO with Length Penalty.}

Finally, we apply GRPO-based reinforcement learning, the reward is computed in two steps. 
We first assign a correctness reward $V(x,y)\in\{0,1\}$ based on whether the final answer is correct. 
We then apply length regularization only to correct trajectories:
\[
R(x,y)=V(x,y)-\lambda\,\mathbf{1}_{\{V(x,y)=1\}}\,R_{\text{length}}(x,y),
\]
To compute the length penalty, we define $L(y)$ as the number of reasoning tokens and the shortest correct trajectory length among the sampled candidates is
\[
L^\star(x)=\min_{y:\,V(x,y)=1} L(y).
\]
We measure the normalized excess length as
\[
\delta(x,y)=\frac{[L(y)-L^\star(x)-\Delta]_+}{L^\star(x)+\Delta},
\]
and define the length regularizer as
\[
R_{\text{length}}(x,y)=\delta(x,y)^\gamma,
\]
where $[z]_+=\max(z,0)$, $\Delta$ is a small tolerance margin, and $\gamma\ge 1$ controls the sharpness of the penalty. 
In this formulation, trajectories whose length is close to the shortest correct one incur almost no penalty, while substantially longer correct trajectories receive increasingly larger penalties, encouraging concise yet accurate reasoning.

\section{Experiments}

\begin{table*}[ht]
\centering
\resizebox{\textwidth}{!}{
\small
\setlength{\tabcolsep}{3.5pt}
\renewcommand{\arraystretch}{1.15}
\begin{tabular}{lcccccccccccc}
\toprule
\multirow{2}{*}{\textbf{Method}} &
\multicolumn{2}{c}{\textbf{AIME24}} &
\multicolumn{2}{c}{\textbf{AIME25}} &
\multicolumn{2}{c}{\textbf{AMC23}} &
\multicolumn{2}{c}{\textbf{OlympiadBench}} &
\multicolumn{2}{c}{\textbf{MATH500}} &
\multicolumn{2}{c}{\textbf{Average}} \\
\cmidrule(lr){2-3}
\cmidrule(lr){4-5}
\cmidrule(lr){6-7}
\cmidrule(lr){8-9}
\cmidrule(lr){10-11}
\cmidrule(lr){12-13}
 & \textbf{Acc$\uparrow$} & \textbf{Len$\downarrow$}
 & \textbf{Acc$\uparrow$} & \textbf{Len$\downarrow$}
 & \textbf{Acc$\uparrow$} & \textbf{Len$\downarrow$}
 & \textbf{Acc$\uparrow$} & \textbf{Len$\downarrow$}
 & \textbf{Acc$\uparrow$} & \textbf{Len$\downarrow$}
 & \textbf{Acc$\uparrow$} & \textbf{Len$\downarrow$} \\
\midrule

\rowcolor[HTML]{F3F8FF}
\multicolumn{13}{l}{\textbf{Open-Source R1-Style Models}} \\

Skywork-OR1-7B\cite{he2025skywork} & 37.08 & 11890 & 24.67 & 12831 & 73.50 & 5241 & 51.41 & 6075 & 85.96 & 3753 & 54.52 & 7958 \\
OREAL-7B\cite{lyu2025exploring} & 38.75 & 10907 & 28.75 & 11309 & 81.25 & 4510 & 56.44 & 5164 & 88.83 & 3023 & 58.80 & 6983 \\
AReaL-boba-RL-7B\cite{mei2025realefficientrlhftraining} & 36.67 & 11407 & 25.42 & 12365 & 71.56 & 5337 & 52.91 & 6037 & 87.12 & 3904 & 54.74 & 7810 \\
Light-R1-DS-7B\cite{lightr1proj} & 39.67 & 12036 & 31.00 & 12498 & 78.25 & 6073 & 54.74 & 5306 & 87.58 & 3020 & 58.25 & 7787 \\

\midrule
\rowcolor[HTML]{E1EDFF}
\multicolumn{13}{l}{\textbf{DeepSeek-R1-Distill-Qwen-1.5B}} \\

Base\cite{deepseekr1} & \textbf{30.83} & 11755 & 22.92 & 11797 & 63.12 & 5205 & 43.89 & 5632 & 72.65 & 2822 & 46.68 & 7442 \\
O1-Pruner$^{*}$\cite{luo2025o1prunerlengthharmonizingfinetuningo1like} & 20.00 & 12680 & 18.75 & 12953 & 59.38 & 5798 & 38.56 & 6645 & 66.81 & 3291 & 40.70 & 8273 \\
TokenSkip$^{*}$\cite{xia-etal-2025-tokenskip} & 18.33 & 12635 & 17.92 & 12630 & 60.62 & 5452 & 38.63 & 6692 & 72.31 & 3139 & 41.56 & 8110 \\
EfficientReasoning\cite{arora2025traininglanguagemodelsreason} & 24.17 & 9380 & 20.00 & 9383 & 68.75 & 3475 & 49.48 & 4032 & \textbf{81.50} & 1600 & 48.78 & 5574 \\
AdaptThink\cite{zhang-etal-2025-adaptthink} & 22.08 & \textbf{5129} & 17.50 & \textbf{3854} & 65.31 & \textbf{2039} & 46.67 & \textbf{2019} & 78.90 & \textbf{987} & 46.09 & \textbf{2806} \\
\rowcolor[HTML]{FFF8DC}
Ours & 26.67 & 8278 & \textbf{23.33} & 7739 & \textbf{69.38} & 2774 & \textbf{49.78} & 3221 & 80.40 & 1798 & \textbf{49.91} & 4762 \\

\midrule
\rowcolor[HTML]{EBF3FF}
\multicolumn{13}{l}{\textbf{DeepSeek-R1-Distill-Qwen-7B}} \\

Base\cite{deepseekr1} & 42.67 & 12723 & 29.00 & 12779 & 81.00 & 6849 & 56.77 & 5252 & \textbf{89.16} & 3065 & 59.72 & 8134 \\
O1-Pruner$^{*}$\cite{luo2025o1prunerlengthharmonizingfinetuningo1like} & 36.67 & 11191 & 27.50 & 11475 & 75.62 & 5128 & 54.15 & 5695 & 76.62 & 2812 & 54.11 & 7260 \\
TokenSkip$^{*}$\cite{xia-etal-2025-tokenskip} & 39.58 & 10841 & 30.83 & 11259 & 79.06 & 4811 & 52.94 & 5795 & 77.56 & 2115 & 55.99 & 6964 \\
EfficientReasoning\cite{arora2025traininglanguagemodelsreason} & 43.75 & 9568 & 30.83 & 9293 & \textbf{83.44} & 3401 & 56.52 & 3920 & 82.25 & \textbf{1392} & 59.36 & 5515 \\
AdaptThink\cite{zhang-etal-2025-adaptthink} & \textbf{44.00} & 9424 & 30.00 & 10290 & 80.75 & 3531 & 58.76 & 5011 & 87.23 & 1603 & 60.15 & 5972 \\
\rowcolor[HTML]{FFF8DC}
Ours & 42.08 & \textbf{7244} & \textbf{31.67} & \textbf{6977} & 82.34 & \textbf{3211} & \textbf{59.85} & \textbf{3786} & 88.80 & 2080 & \textbf{60.95} & \textbf{4660} \\

\bottomrule
\end{tabular}
}
\caption{
Performance comparison across multiple math reasoning benchmarks (Accuracy $\uparrow$ and Average Length $\downarrow$).\\
{\footnotesize $^{*}$ We adapt the original implementation of O1-Pruner and TokenSkip to the DeepSeek-R1-Distill-Qwen models.}
}
\label{tab:math_models_comparison}
\end{table*}

\subsection{Experiment Setup}
\noindent{\textbf{Benchmarks and Metrics:}} We evaluate our method on five widely used mathematical reasoning benchmarks covering different difficulty levels. 
AIME24 and AIME25 are derived from the American Invitational Mathematics Examinations, representing competition-level reasoning problems~\cite{aime2024,aime2025}. 
AMC23 consists of questions from American Mathematics Competitions, reflecting moderately challenging contest problems~\cite{amc23}. 
MATH500 is a curated subset of 500 problems from the MATH benchmark, spanning algebra, number theory, geometry, and probability~\cite{math500}. 
OlympiadBench further tests advanced mathematical and scientific reasoning with bilingual Olympiad-level problems~\cite{olympiadbench}. For all the datasets, we sample 10 solutions for each problem, we then calculate the average accuracy of the 10 solutions using math-verify and report the average number of generated tokens across all samples.

\noindent{\textbf{Baselines:}} 
We compare our method with several efficient reasoning methods. (1) \textbf{O1-Pruner}\cite{luo2025o1prunerlengthharmonizingfinetuningo1like}: a length-harmonizing fine-tuning approach that reduces long CoT traces while preserving performance. 
(2) \textbf{TokenSkip}\cite{xia-etal-2025-tokenskip}: a controllable CoT compression method that estimates token-level importance and removes low-utility tokens to shorten reasoning traces. 
(3) \textbf{EfficientReasoning}\cite{arora2025traininglanguagemodelsreason}: an RL-based objective that favors correct yet concise reasoning, improving efficiency while maintaining accuracy. 
(4) \textbf{AdaptThink}\cite{zhang-etal-2025-adaptthink}: an RL method that trains the model to adaptively decide whether to generate an explicit reasoning trace or respond directly based on input difficulty.
Additionally, we also compare against several open-source R1-like reasoning models, including \textbf{Skywork-OR1-7B}\cite{he2025skywork}, \textbf{OREAL-7B}\cite{lyu2025exploring}, \textbf{AReaL-base-R1-7B}\cite{mei2025realefficientrlhftraining}, and \textbf{Light-R1-DS-7B}\cite{lightr1proj}.

\noindent\textbf{Training Details:} 
We conduct all training on DeepSeek-R1-Distill-Qwen-1.5B and DeepSeek-R1-Distill-Qwen-7B\cite{deepseekr1}. We start with supervised SFT on the Light-R1~\cite{lightr1proj} dataset. We prune all CoT traces with our graph-based method to remove redundant reflection nodes while preserving the core reasoning structure, and then fine-tune the model to imitate these concise reasoning traces. Next, we sample problems from AIME (pre-2024) and AMC (pre-2023)\cite{numina_math_datasets} and use the SFT model to generate responses. For each problem, we construct preference pairs by ranking sampled correct trajectories by our redundancy score, treating low-redundancy trajectories as preferred and high-redundancy ones as dispreferred, and train a DPO model that favors shorter yet effective reasoning. Finally, we perform GRPO with length penalty on the dapo-17k~\cite{yu2025dapoopensourcellmreinforcement} dataset with verifiable rewards and length penalty to further improve correctness and robustness. All experiments are conducted on a single compute node with $4\times$ NVIDIA A800 GPUs.

\vspace{-3mm}

\begin{figure*}[t]
  \centering
  \begin{subfigure}[t]{0.49\textwidth}
    \centering
    \includegraphics[width=\linewidth]{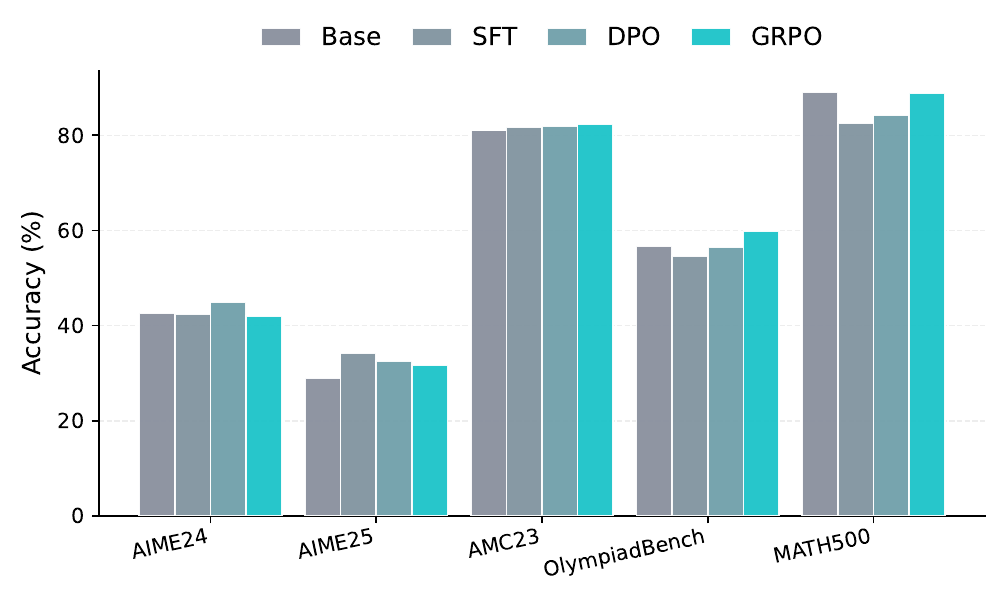}
    \caption{}
    \label{fig:ablation-a}
  \end{subfigure}
  \hfill
  \begin{subfigure}[t]{0.49\textwidth}
    \centering
    \includegraphics[width=\linewidth]{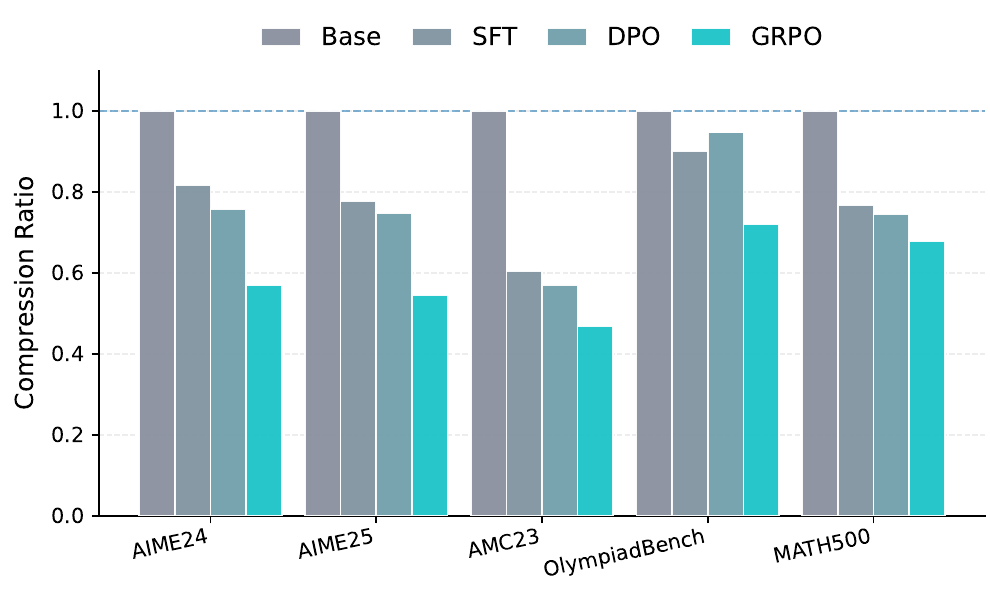}
    \caption{}
    \label{fig:ablation-b}
  \end{subfigure}

  \caption{\textbf{Stage-wise ablation across five benchmarks.} (a) Accuracy (\%). (b) Relative reasoning length measured by the average number of generated tokens, normalized to the \textsc{Base} setting (token ratio; \textsc{Base}=1.0, lower is better). The compared configurations follow a cumulative training recipe: starting from \textsc{Base}, we sequentially add \textsc{SFT}, then \textsc{DPO}, and finally \textsc{GRPO}.}
  \label{fig:ablation}
\end{figure*}

\subsection{Main Results}

We compare our method against four widely used efficiency-oriented baselines at two model scales, with results reported in Table~\ref{tab:math_models_comparison}. Overall, our approach achieves the best accuracy--efficiency trade-off: it attains the highest average accuracy while producing substantially shorter reasoning traces. On DeepSeek-R1-Distill-Qwen-7B, our method improves the average accuracy from 59.72 to 60.95 while reducing the average reasoning length from 8134 to 4660 tokens (42.7\% reduction). The gains are most evident on difficult benchmarks: for AIME25, accuracy increases from 29.00\% to 31.67\% while the reasoning length drops from 12779 to 6977 tokens; for OlympiadBench, accuracy increases from 56.77\% to 59.85\% with shorter traces (5252 $\rightarrow$ 3786). On DeepSeek-R1-Distill-Qwen-1.5B, our method also improves the average accuracy from 46.68 to 49.91 and reduces the average length from 7442 to 4762 tokens (a 36\% reduction), with notable gains on AMC23 (63.12\% $\rightarrow$ 69.38\%) and MATH500 (72.65\% $\rightarrow$ 80.40\%). These results hold across datasets and model sizes, indicating that our method scales well and consistently reduces redundant computation without sacrificing overall performance.

\subsection{Ablation Study}
We conduct a stage-wise ablation to disentangle the contributions of each component in our training recipe to both accuracy and efficiency. Starting from the base policy (DS-7B), we progressively add SFT, DPO, and GRPO. 
Figure~\ref{fig:ablation} summarizes the results across five benchmarks. We report (a) accuracy and (b) the average number of generated tokens normalized to the Base setting (token ratio; Base$=1.0$, lower is better). 
This protocol separates performance improvements from changes in reasoning length, allowing us to assess whether generation cost can be reduced without sacrificing accuracy.

\section{Analysis of Graph-based CoT Pruning}
\label{sec:analysis}

In this section, we provide a detailed analysis of our graph-based CoT pruning framework from four perspectives: (i) dataset- and graph-level statistics, (ii) whether pruning preserves essential reasoning, and (iii) changes in model behavior before and after training.

\subsection{Dataset and Graph Statistics}

We first examine how graph-based pruning reshapes the structure of CoT supervision data. We report the average number of nodes per reasoning graph, the number of reflection nodes, the number of redundant reflection nodes identified by our pruning algorithm, the average CoT length in tokens, the proportion of main-path nodes, as well as the total number of training examples and the data synthesis cost. Table~\ref{tab:graph_stats} shows the statistics that graph-based pruning removes a large fraction of redundant reflection nodes while keeping the main reasoning path relatively intact, substantially reducing supervision length without discarding most of the essential steps and incurring only a low data synthesis cost.

\begin{table}[h]
  \centering
  \normalsize
  \setlength{\tabcolsep}{4pt}      
  \renewcommand{\arraystretch}{1.1} 
  \begin{tabular}{lcc}
    \toprule
    \textbf{Statistic} & \textbf{Full CoT} & \textbf{Pruned CoT} \\
    \midrule
    Total Samples                & 3335   & 3335   \\
    Avg.\ Nodes              & 27.8   & 15.6   \\
    Avg.\ Review Nodes       & 16.8  & 4.5   \\
    Avg.\ Tokens      & 6468  & 4439   \\
    Total Cost          & --     & \$20     \\
    \bottomrule
  \end{tabular}
  \caption{
    Dataset and graph statistics before and after pruning.
  }
  \label{tab:graph_stats}
\end{table}

\begin{figure*}[t]
    \centering
    \includegraphics[width=\textwidth]{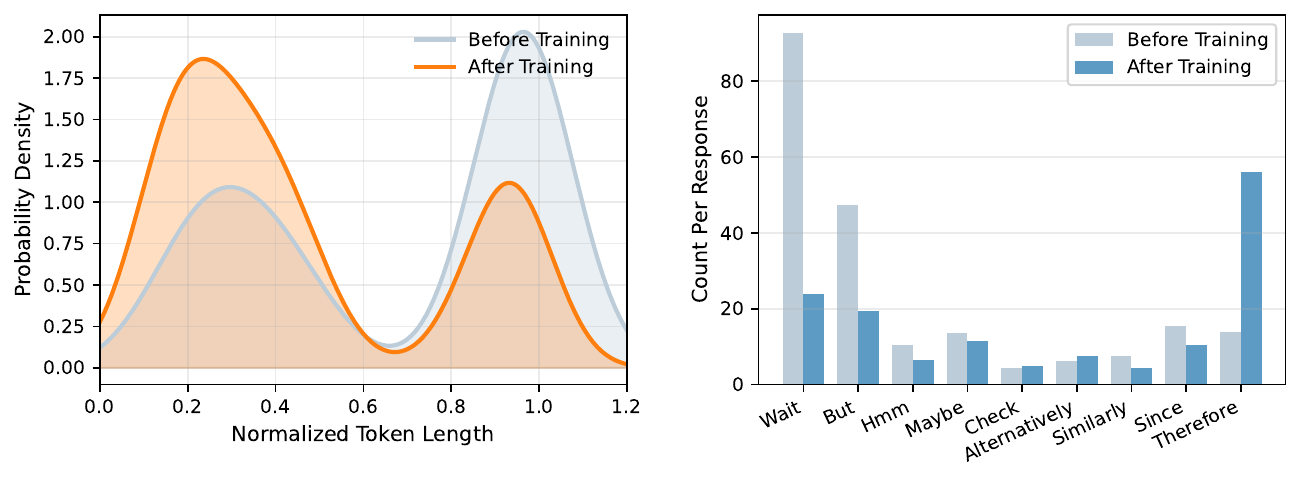}
    \caption{Changes in reasoning length and reasoning token usage before and after training on AIME24.}
    \label{fig:len_and_keywords}
\end{figure*}

\subsection{Does Pruning Preserve Essential Reasoning?}

To investigate whether pruning inadvertently removes crucial reasoning steps, we established three distinct experimental settings: (1) \textbf{Full-CoT}: the original, complete reasoning traces, (2) \textbf{Graph-Pruned}: traces processed via our graph-based pruning, and (3) \textbf{Len-Trunc}: traces truncated from the beginning to match the token length of the Graph-Pruned variants. We randomly sampled 1,000 examples from the training set. For each example, we employed \texttt{DeepSeek-R1-Distill-Qwen-7B} to generate answers conditioned on these different CoT types, performing $N=8$ generation passes per question to assess robustness.

We evaluated the model using two metrics: \textbf{Accuracy}, representing the percentage of questions answered correctly, and \textbf{Consistency}, which measures the degree of consensus among the 8 generated answers. Consistency is defined as $C = \sum (n_y / N)^2$, where $n_y$ denotes the frequency of a specific answer $y$ across the $N$ generations; a higher score indicates the model reliably converges on the same output. As shown in Table~\ref{tab:cot_follow_eval}, \textbf{Full-CoT} sets a high baseline with 98.95\% accuracy and 99.60\% consistency. \textbf{Graph-Pruned} maintains high reliability, achieving 93.70\% accuracy and 90.69\% consistency. In contrast, \textbf{Len-Trunc} suffers a significant drop to 73.60\% accuracy and 69.10\% consistency. This stark difference indicates that while naive length truncation disrupts the logical flow, leading to divergent and incorrect answers, our graph-based pruning successfully preserves the core reasoning structure necessary for stable and accurate generation.

\begin{table}[t]
\centering
\setlength{\tabcolsep}{7pt}
\renewcommand{\arraystretch}{1.15}

\resizebox{\columnwidth}{!}{%
\begin{tabular}{lcc}
\toprule
\textbf{CoT Variant} & \textbf{Accuracy ($\uparrow$)} & \textbf{Consistency ($\uparrow$)} \\
\midrule
Full-CoT     & 98.95 & 99.60 \\
Graph-Pruned & 93.70 & 90.69 \\
Len-Trunc    & 73.60 & 69.10 \\
\bottomrule
\end{tabular}%
}

\caption{Comparison of accuracy and consistency across different Chain-of-Thought (CoT) pruning methods.}
\label{tab:cot_follow_eval}
\end{table}




\subsection{Impact on Training and Model Behavior}

\paragraph{Reasoning length and reflection tokens.}
We analyze how the model’s test-time reasoning behavior changes before and after training. 
We first examine the distribution of reasoning length, measured by the number of generated tokens and normalized across samples. 
As shown in Figure~\ref{fig:len_and_keywords}, the density curves indicate that, after training, the model produces noticeably shorter reasoning trajectories, with a clear suppression of the long-tail region corresponding to excessively long responses. 

We then analyze the frequency of representative reasoning tokens (e.g., ``wait'', ``but'', ``hmm'', ``maybe'', and ``check'') on AIME24 by counting their occurrences per response. As shown in the bar plot, these reflection-oriented tokens consistently decrease after training, indicating reduced reflective behaviors. In contrast, progress-oriented connectives such as ``therefore'' become substantially more frequent, suggesting a shift from reflective verbosity toward more direct, decision-driven reasoning.

\paragraph{Qualitative case studies.}
Finally, we present qualitative case studies to illustrate how graph-based pruning reshapes the reasoning process.
For selected problems, we visualize the original CoT graph and the pruned graph, highlighting nodes identified as redundant reflections.
We also show the corresponding natural-language CoT segments with removed pieces struck out or shaded.
In most cases, pruning eliminates repeated self-checks and digressions while keeping the core derivation intact, leading to cleaner and more stable reasoning trajectories.
An example is shown in Figure~\ref{fig:case_study} for reference.

\section{Conclusion}

We presented a graph-based approach for improving CoT efficiency by identifying and pruning redundant reflection steps that do not contribute to the main reasoning path. By converting linear CoTs into structured graphs, our method localizes and removes low-importance review nodes while preserving essential logic, yielding more concise CoTs without sacrificing accuracy. Our findings show that structural modeling of reasoning offers a promising direction for systematically reducing overthinking and enabling more efficient LLM reasoning.

\section*{Limitations}

Our approach requires constructing reasoning graphs with a strong teacher model, which introduces preprocessing cost and may limit scalability. The progress–review labeling is coarse and may overlook fine-grained reasoning nuances. Additionally, while effective for mathematical reasoning, it remains unclear how well the method generalizes to more open-ended domains. Future work may explore lighter-weight graph construction, richer semantic labels, and broader domain evaluations.


\nocite{*}
\bibliography{custom}

\appendix

\clearpage
\section{Prompts}
\label{app:graph_prompts}

This section describes the instruction prompt used to construct a graph based on the model's original CoT. Given a reasoning step and a partial graph, the model updates the graph by either inserting a new node or merging the step into an existing one.

Each node represents an abstract reasoning step and is labeled as either progress, which advances the reasoning process, or review, which captures reflective behavior. Reflective steps are prohibited from being merged into progress nodes.

The prompt enforces fixed node identifiers, directed dependency edges, and a dedicated \textit{final answer} node, with all updates returned in a structured JSON format for deterministic parsing. The full prompt is shown in Figure~\ref{fig:graph_prompt}.

\section{Special Tokens to Split CoT to Chunks}
\label{app:split_tokens}
We use a set of special tokens to split a long reasoning trajectory into multiple step-level chunks.
\begin{lstlisting}[basicstyle=\ttfamily\small,breaklines=true,breakatwhitespace=true]
split_tokens = [ "Wait", "Alternatively", "Another angle", "Another approach", "But wait", "Hold on", "Hmm", "Maybe", "Looking back", "Okay", "Let me", "First", "Then", "Alright", "Compute", "Correct", "Good", "Got it", "I don't see any errors", "I think", "Let me double-check", "Let's see", "Now", "Remember", "Seems solid", "Similarly", "So", "Starting", "That's correct", "That seems right", "Therefore", "Thus" ]
\end{lstlisting}

\section{Node-level Evaluation of Graph Construction}
\label{app:graph_eval}

To evaluate the accuracy of converting linear CoT into graph-structured representations, we conduct a human evaluation on a randomly sampled set of 100 graph nodes.
Each node is assessed along two dimensions.

First, we evaluate node type correctness.
Nodes are classified as either \emph{progress} or \emph{review}, and annotators judge whether the predicted type matches the node’s functional role in the original reasoning process.
We report per-class precision, recall, and F1 scores for both node types.

Second, we evaluate step atomicity.
A node is considered atomic if it corresponds to a single, semantically independent reasoning step, without mixing multiple operations.
We report the atomicity valid rate as a measure of structural quality.

We further define a node as valid only if it satisfies both criteria.
Table~\ref{tab:type_atomicity} summarizes the results.
The model achieves high accuracy in node type classification and a strong atomicity valid rate, indicating that the constructed graphs are both semantically faithful and structurally well-formed.

\begin{table}[H]
\centering
\small
\setlength{\tabcolsep}{8pt}
\renewcommand{\arraystretch}{1.15}
\begin{tabular}{lccc}
\toprule
\textbf{Node Type} 
& \textbf{Precision} 
& \textbf{Recall} 
& \textbf{F1} \\
\midrule
Review    & 0.9048 & 0.9661 & 0.9344 \\
Progress  & 0.8947 & 0.8500 & 0.8718 \\
\midrule
\textbf{Atomicity Valid (\%)} & \multicolumn{3}{c}{\textbf{85.29}} \\
\bottomrule
\end{tabular}
\caption{Type classification performance and step atomicity validity of constructed graph nodes.}
\label{tab:type_atomicity}
\end{table}

\section{Implementation Details}
\label{app:impl}

\subsection{SFT Training Settings}
We perform SFT using the LLaMA-Factory framework with LoRA-based parameter-efficient tuning. LoRA adapters are applied to all attention and linear layers. The hyper-parameters are summarized in Table~\ref{tab:sft_hparams}.

\begin{table}[H]
\centering
\normalsize
\begin{tabular}{lc}
\toprule
\textbf{Name} & \textbf{Value} \\
\midrule
Epochs              & 8 \\
Global batch size   & 64 \\
Max sequence length & 8192 \\
Optimizer           & AdamW \\
Learning rate       & $1\times10^{-5}$ \\
LoRA rank           & 32 \\
\bottomrule
\end{tabular}
\caption{Hyper-parameters for SFT training.}
\label{tab:sft_hparams}
\end{table}

\subsection{DPO Training Settings}
We perform DPO using the same training framework as SFT. The hyper-parameters are summarized in Table~\ref{tab:dpo_hparams}.

\begin{table}[H]
\centering
\normalsize
\begin{tabular}{lc}
\toprule
\textbf{Name} & \textbf{Value} \\
\midrule
Epochs                 & 5 \\
Global batch size      & 64 \\
Max sequence length    & 8192 \\
Optimizer              & AdamW \\
Learning rate          & $1\times10^{-7}$ \\
\bottomrule
\end{tabular}
\caption{Hyper-parameters for DPO training.}
\label{tab:dpo_hparams}
\end{table}

\subsection{GRPO Training Settings}
We conduct GRPO with length penalty using the verl framework. The hyper-parameters are summarized in Table~\ref{tab:grpo_hparams}.

\begin{table}[H]
\centering
\normalsize
\begin{tabular}{lc}
\toprule
\textbf{Name} & \textbf{Value} \\
\midrule
KL coefficient $\beta$ & $1\times10^{-3}$ \\
Rollouts per input     & 8 \\
Sampling temperature   & 1.0 \\
Max response length    & 12000 \\
Global batch size      & 64 \\
Optimizer              & AdamW \\
Learning rate          & $1\times10^{-6}$ \\
Training steps         & 220 \\
\bottomrule
\end{tabular}
\caption{Hyper-parameters for GRPO training.}
\label{tab:grpo_hparams}
\end{table}

\section{RL Training Dynamics}
\label{app:rl_dynamics}

\begin{figure}[H]
    \centering

    \begin{subfigure}[t]{0.48\linewidth}
        \centering
        \includegraphics[width=\linewidth]{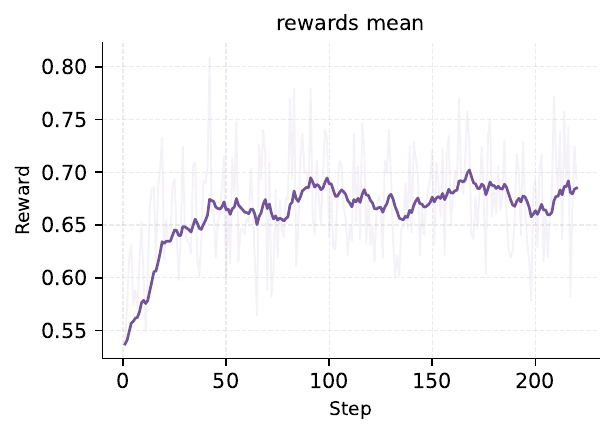}
        \label{fig:reward_curve_7b}
    \end{subfigure}\hfill
    \begin{subfigure}[t]{0.48\linewidth}
        \centering
        \includegraphics[width=\linewidth]{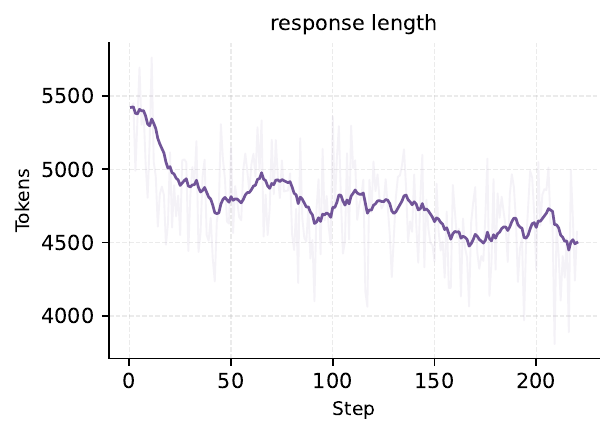}
        \label{fig:length_curve_7b}
    \end{subfigure}

    \vspace{-2mm}
    {\footnotesize (a)}

    \vspace{2mm}

    \begin{subfigure}[t]{0.48\linewidth}
        \centering
        \includegraphics[width=\linewidth]{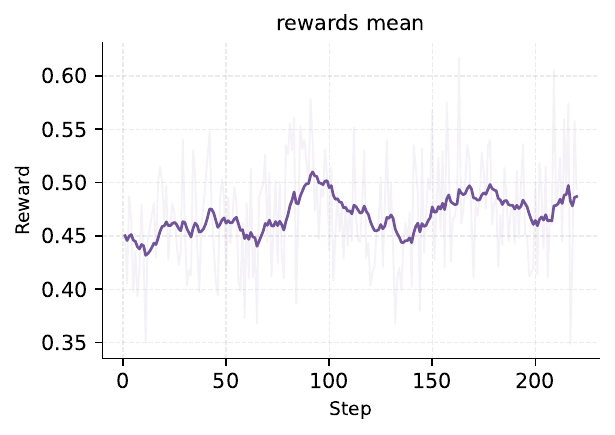}
        \label{fig:reward_curve_1p5b}
    \end{subfigure}\hfill
    \begin{subfigure}[t]{0.48\linewidth}
        \centering
        \includegraphics[width=\linewidth]{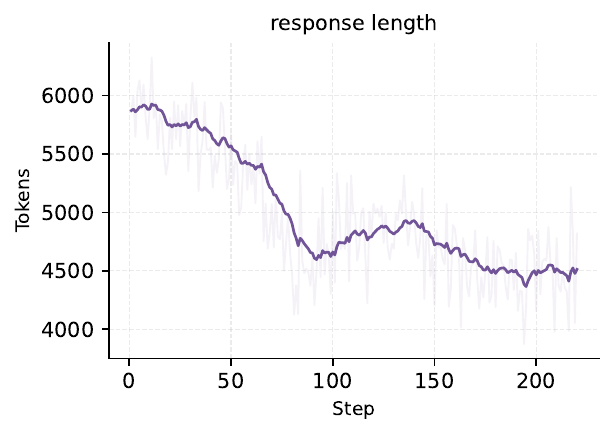}
        \label{fig:length_curve_1p5b}
    \end{subfigure}

    \vspace{-2mm}
    {\footnotesize (b)}

    \vspace{-2mm}
    \caption{RL training dynamics for different model scales. (a) 7B; (b) 1.5B. Left: mean reward. Right: response length (tokens).}

    \label{fig:rl_dynamics}
\end{figure}

Figure~\ref{fig:rl_dynamics} presents the training dynamics of the model during reinforcement learning. We visualize the evolution of the reward signal and the average response length across training steps. To reduce high-frequency noise inherent to reinforcement learning, we apply exponential moving average (EMA) smoothing to the raw curves.

As shown in the figure, the reward exhibits an overall increasing trend despite noticeable fluctuations, which is typical for policy optimization with sparse or delayed rewards. At the same time, the response length does not grow monotonically with reward improvement, indicating that higher rewards are not solely achieved by generating longer responses. Instead, the model gradually learns more effective reasoning strategies under the given reward signal.

\section{Overall Training Algorithm}

Algorithm~\ref{alg:three_stage} presents the pseudocode of our training algorithm, detailing how the policy is optimized across SFT, DPO, and GRPO-based RL.

\begin{algorithm}
\small
\caption{Three-Stage Training with Graph Construction and Pruning}
\label{alg:three_stage}
\KwIn{Problems $\{q\}$, raw CoT $r=[c_1,\ldots,c_{n}]$ and answer $a$ for each $q$, base policy $M_\theta$}
\KwOut{Final policy $M_\theta$}

\BlankLine
\textbf{(A) Construct pruned traces for SFT.}\;
\ForEach{problem $q$ with CoT $r=[c_1,\ldots,c_{n}]$}{
    $G_0\leftarrow(\varnothing,\varnothing,\varnothing)$\;
    \For{$i=1$ \KwTo $n$}{
        $o_i \sim f_L(G_{i-1},c_i)$, \quad $o_i\in\{\texttt{insert},\texttt{merge}\}$\;
        Update $G_i$ by applying $o_i$ (create/merge node $v_i=(s_i,l_i)$ and add dependency edges)\;
    }
    $\tilde G \leftarrow \textsc{Prune}(G_{n}; m,k)$\;
    $\tilde r \leftarrow \textsc{Relinearize}(\tilde G)$\;
    Add $(q,\tilde r,a)$ to $\mathcal{D}_{\text{SFT}}$\;
}

\BlankLine
\textbf{(B) Cold-start SFT.}\;
Train $M_\theta$ on $\mathcal{D}_{\text{SFT}}$ by minimizing $\mathcal{L}_{\text{SFT}}(\theta)$ to obtain $M_{\text{SFT}}$\;

\BlankLine
\textbf{(C) DPO via rollout and preference pairing.}\;
Initialize $M_\theta \leftarrow M_{\text{SFT}}$\;
\ForEach{problem $x$}{
    Sample rollouts $\mathcal{Y}=\{y^{(k)}\}$ from $\pi_\theta(\cdot\mid x)$\;
    Keep correct set $\mathcal{Y}_{\text{ok}}=\{y\in\mathcal{Y}\mid V(x,y)=1\}$\;
    Score redundancy for each $y\in\mathcal{Y}_{\text{ok}}$\;
    Form preference pairs $(y^+,y^-)$ with lower-vs-higher redundancy\;
    Update $M_\theta$ by minimizing $\mathcal{L}_{\text{DPO}}(\theta)$ on collected pairs to obtain $M_{\text{DPO}}$\;
}

\BlankLine
\textbf{(D) GRPO with length penalty.}\;
Initialize $M_\theta \leftarrow M_{\text{DPO}}$\;
\ForEach{problem $x$}{
    Sample trajectories $\mathcal{Y}\sim \pi_\theta(\cdot\mid x)$\;
    Compute shortest correct length $L^\star(x)$ in $\mathcal{Y}$\;
    \ForEach{$y\in\mathcal{Y}$}{
        Compute accuracy reward and length reward\;
    }
    Update $M_\theta$ with GRPO using reward $R$\;
}
\end{algorithm}

\begin{figure*}[t]
\centering
\begin{spacing}{0.95}
\begin{promptbox}[Prompt for Constructing Graph]
\small
\setlist[enumerate]{nosep, topsep=2pt, partopsep=0pt, leftmargin=1.5em}
\setlist[itemize]{nosep, topsep=2pt, partopsep=0pt, leftmargin=1.5em}
\renewcommand{\medskip}{\vspace{2pt}}

You are a \textbf{chain-of-thought graph structure analysis and update module}.\\
Given an existing \texttt{graph} and the current text segment \texttt{current\_step}, you must incorporate \texttt{current\_step} into the graph by choosing \textbf{exactly one} operation: \textbf{Insert} or \textbf{Merge}, and output \textbf{strict JSON}.

\medskip
\textbf{1. Inputs}
\begin{itemize}
  \item \texttt{graph}: an existing partial reasoning graph (Mermaid code).
  \item \texttt{current\_step}: the current reasoning text segment (continuous content from the CoT).
\end{itemize}

\medskip
\textbf{2. Node Definition (Reasoning Unit)}
\begin{itemize}
  \item A node represents an \textbf{abstract reasoning unit} with:
  \begin{itemize}
    \item \textbf{Semantic completeness}: a clear intent and its reasoning product (e.g., introducing a constraint, deriving a conclusion, setting a sub-goal, establishing a framework).
    \item \textbf{Abstraction}: do \emph{not} record low-level arithmetic/symbolic manipulation.
    \item \textbf{Dependability}: its product can be referenced by later reasoning and creates dependencies.
  \end{itemize}
  \item \textbf{Forbidden}: purely operational steps (substitution, expansion, simplification, step-by-step calculations) cannot be standalone nodes.
\end{itemize}

\medskip
\textbf{3. Independence Criteria}
Treat \texttt{current\_step} as a new reasoning unit (prefer \textbf{Insert}) if it satisfies any of:
\begin{itemize}
  \item \textbf{Goal introduction}: introduces a new intermediate goal/subproblem.
  \item \textbf{Product generation}: yields a key conclusion/property/constraint/equivalence used later.
  \item \textbf{Method switch}: changes the reasoning strategy or framework.
  \item \textbf{Structural advancement}: adds structure (case split, construction, invariant/lemma framework).
  \item \textbf{Branch initiation}: starts a new attempt path (even if it fails); branch from still-valid ancestors.
\end{itemize}
If it only continues the same goal with minor details/restatement and produces no new product/structure, prefer \textbf{Merge}.

\medskip
\textbf{5. Update Operations}
\begin{itemize}
  \item \textbf{Merge}: Allowed only if \texttt{current\_step} can be integrated into exactly one existing node while keeping it \textbf{abstract} and \textbf{single-purpose}.
  \textbf{Hard constraints}: Review content \textbf{cannot} be merged into a \textbf{progress} node; Do not merge if it causes one node to mix ``advance'' and ``reflect'' as major functions.
  Must specify \texttt{target\_node} and \texttt{updated\_node\_description}.

  \item \textbf{Insert}: Required if \texttt{current\_step} meets independence criteria, introduces a new branch/framework, or Merge would harm abstraction/readability.
  Must create a new node and add necessary dependency edges.
\end{itemize}

\medskip
\textbf{6. Edge Construction Rules}
\begin{itemize}
  \item \textbf{Dependency principle}: if node B uses products from node A, add edge $A \rightarrow B$ with a clear dependency label.
  \item \textbf{Branch origin principle}: new attempts must branch from \textbf{still-valid ancestor products}, not from negated/dead-end nodes.
  \item \textbf{Ordering constraints}:
  \begin{itemize}
    \item Node IDs increase lexicographically: A--Z, AA--AZ, BA--BZ, \ldots
    \item Edges must go from lexicographically smaller IDs to larger IDs.
    \item The final node must be named \textit{final answer}.
  \end{itemize}
\end{itemize}

\medskip
\textbf{7. Output Format (Strict JSON)}
\begin{verbatim}
{
  "decision": "Insert or Merge",
  "target_node": "(If Merge, the node ID to merge into; if Insert, leave empty)",
  "new_node": {
    "id": "(If Insert, a unique ID; if Merge, leave empty)",
    "description": "(If Insert, a concise description of the new node; if Merge, leave empty)",
    "type": "progress or review"
  },
  "edges": [
    {
      "from": "source node ID",
      "to": "target node ID",
      "label": "meaning of the edge"
    }
  ],
  "updated_node_description": "(If Merge, the new description; if Insert, leave empty)"
}
\end{verbatim}

\textbf{Your Turn:}\\
Existing Graph: \texttt{\{\{ graph \}\}}\\
Current Step: \texttt{\{\{ current\_step \}\}}\\
Your Response:
\end{promptbox}
\end{spacing}
\caption{Instruction for updating a graph-structured chain-of-thought}
\label{fig:graph_prompt}
\end{figure*}

\begin{figure*}[ht]
    \centering
    \includegraphics[width=\textwidth]{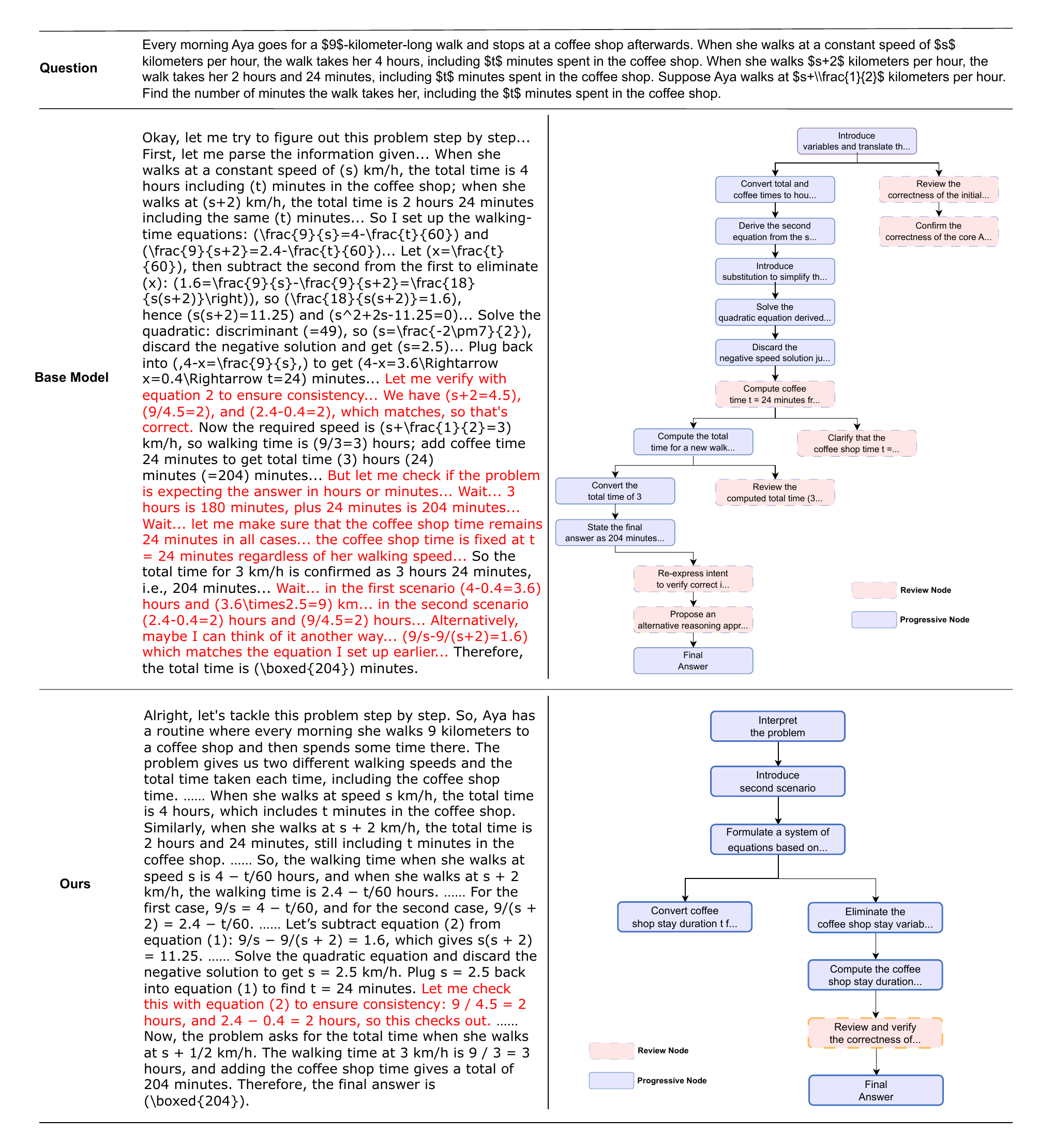}
    \caption{Qualitative comparison of reflection behavior between the base model and our trained model. The left column shows the original CoT, and the right column shows its graph-structured representation. Red-highlighted text indicate reflection-related content. Compared to the base model, our trained model exhibits reduced and more focused reflection.}
    \label{fig:case_study}
\end{figure*}

\end{document}